\documentclass[letterpaper, 10 pt, conference]{ieeeconf}  

\IEEEoverridecommandlockouts                              
\usepackage{booktabs}

\usepackage{cite}
\usepackage{amsmath,amssymb,amsfonts}
\usepackage{algorithmic}
\usepackage{graphicx}
\usepackage{textcomp}
\usepackage{xcolor}
\usepackage{subfig}

\usepackage{textcomp}

\usepackage{fancyhdr}
\usepackage[ruled,vlined]{algorithm2e}

\usepackage{amsmath}
\usepackage{fancyhdr,graphicx,amsmath,amssymb}
\usepackage[ruled,vlined]{algorithm2e}
\usepackage{bm}
\usepackage{xcolor}

\usepackage{url}
\usepackage{hyperref}
\hypersetup{
    colorlinks=true,
    linkcolor=blue,
    filecolor=magenta,      
    }

\usepackage{multirow} 
\let\labelindent\relax
\DeclareUnicodeCharacter{2212}{-}

\def\BibTeX{{\rm B\kern-.05em{\sc i\kern-.025em b}\kern-.08em
    T\kern-.1667em\lower.7ex\hbox{E}\kern-.125emX}}

\usepackage{bm} 
\usepackage{color} 
\usepackage{siunitx}
\usepackage{tikz}
\usetikzlibrary{decorations.pathmorphing} 
\usetikzlibrary{matrix} 
\usetikzlibrary{arrows} 
\usetikzlibrary{calc} 
\usepackage{tikzscale}

\tikzstyle{block} = [draw,rectangle,thick,minimum height=2em,minimum width=2em]
\tikzstyle{sum} = [draw,circle,inner sep=0mm,minimum size=2mm]
\tikzstyle{connector} = [->,thick]
\tikzstyle{line} = [thick]
\tikzstyle{branch} = [circle,inner sep=0pt,minimum size=1mm,fill=black,draw=black]
\tikzstyle{guide} = []
\tikzstyle{snakeline} = [connector, decorate, decoration={pre length=0.2cm,
                         post length=0.2cm, snake, amplitude=.4mm,
                         segment length=2mm},thick, magenta, ->]

\usepackage{etoolbox}
\makeatletter
\patchcmd{\@makecaption}
  {\scshape}
  {}
  {}
  {}
\makeatletter
\patchcmd{\@makecaption}
  {\\}
  {.\ }
  {}
  {}
\makeatother
\def\tablename{Table}
\def\tablename{Table}

\usepackage[compact]{titlesec}
\titlespacing{\section}{2pt}{*+1}{*+1}
\titlespacing{\subsection}{2pt}{*+1}{*+1}
\titlespacing{\subsubsection}{2pt}{*+1}{*+1}

\usepackage{enumitem}
\setlist{nolistsep,leftmargin=*}
\setlist{nolistsep}
\usepackage{amsmath}
\usepackage{fancyhdr,graphicx,amsmath,amssymb}
\usepackage[ruled,vlined]{algorithm2e}
\usepackage{bm}
\usepackage{xcolor}
\usepackage[font=footnotesize]{caption}

\begin{document}
\include{pythonlisting}
\title{APEX: Ambidextrous Dual-Arm Robotic Manipulation Using \\ Collision-Free Generative Diffusion Models}

\author{Apan Dastider, Hao Fang, and Mingjie Lin}

    

\twocolumn[{%
\renewcommand\twocolumn[1][]{#1}%
\maketitle
\begin{center}
    \centering
    \captionsetup{type=figure}
    \includegraphics[width=1\linewidth]{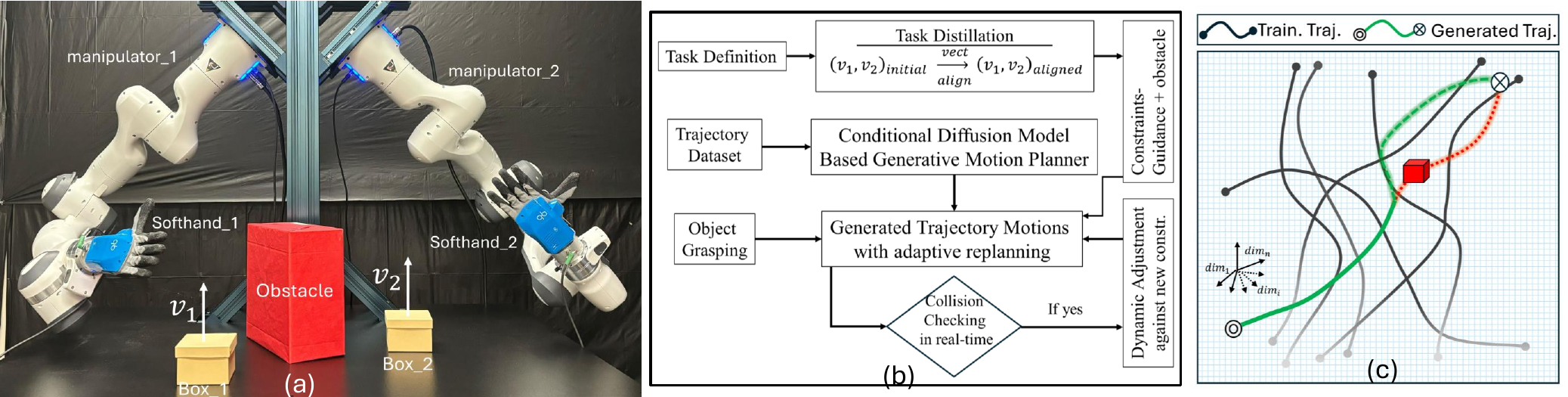}
    \captionof{figure}{a)Dual-arm robotic manipulation platform (b) Overall diagram of our algorithm. (c) Our diffusion models learn the distribution of trajectory using training data (black line) and generate a new planning trajectory (green dashed line). Further,  it adapts the generated trajectory (red dashed line) to avoid the potential obstacles (red box).}
    \label{fig: title}
\end{center}%
}]


\begin{abstract}


Dexterous manipulation, particularly adept coordinating and grasping, constitutes a fundamental and indispensable capability for robots, facilitating the emulation of human-like behaviors. Integrating this capability into robots empowers them to supplement and even supplant humans in undertaking increasingly intricate tasks in both daily life and industrial settings.
Unfortunately, contemporary methodologies encounter serious challenges in devising manipulation trajectories owing to the intricacies of tasks, the expansive robotic manipulation space, and dynamic obstacles. 
We propose a novel approach, APEX, to address all these difficulties by introducing a collision-free latent diffusion model for both robotic motion planning and manipulation.
Firstly, we simplify the complexity of real-life ambidextrous dual-arm robotic manipulation tasks by abstracting them as aligning two vectors. Secondly, we devise latent diffusion models to produce a variety of robotic manipulation trajectories. Furthermore, we integrate obstacle information utilizing a classifier-guidance technique, thereby guaranteeing both the feasibility and safety of the generated manipulation trajectories. Lastly, we validate our proposed algorithm through extensive experiments conducted on the hardware platform of ambidextrous dual-arm robots. Our algorithm consistently generates successful and seamless trajectories across diverse tasks, surpassing conventional robotic motion planning algorithms. These results carry significant implications for the future design of diffusion robots, enhancing their capability to tackle more intricate robotic manipulation tasks with increased efficiency and safety. Complete video demonstrations of our experiments can be found in
\url{https://sites.google.com/view/apex-dual-arm/home}.

\end{abstract}

 
\section{Introduction}           

Ambidextrous dual-arm robotic manipulation requires sophisticated coordination and control to synchronize the movements of both arms and hands seamlessly. Unlike well-studied single-arm robots, ambidextrous robots must coordinate the actions of two separate arms and two hands while navigating complex environments and interacting with dynamic objects. This heightened level of complexity necessitates advanced planning and control strategies to ensure successful task execution and obstacle avoidance. 
However, realizing such ambidextrous dual-arm robotic manipulation in real world is complicated~\cite{Abbas2023}. First, different from the prior 7-DoF robotic arm model, the
ambidextrous dual-arm robotic system allows more joint freedom, which requires
more parameters and faces complex constraints to comprehensively describe the joint configuration space~\cite{zhao_dual_arm}.
Second, the aim of the ambidextrous
dual-arm robotic system is to perform more complicated human-like tasks, such as packaging, stacking, steering liquid in a cup, and etc; which impose more dexterous
manipulation requirements such as smoothness and diversity. Third, the real-time
moving obstacles should be avoided during manipulation, which
requires the algorithm to work under a close-loop fashion~\cite{Dastider_Damon_IROS_2023}. Fourth, it is crucial that motion planning algorithms exhibit computational efficiency to support real-time adaptive control of both robotic arms and hands~\cite{He2022_dual_arm_opt_control}. 
All these challenges collectively impede the advancement of motion planning algorithms in dual-arm robotic manipulation tasks.

\begin{figure*}                                     
    \centering                                             
    \includegraphics[width=\linewidth]{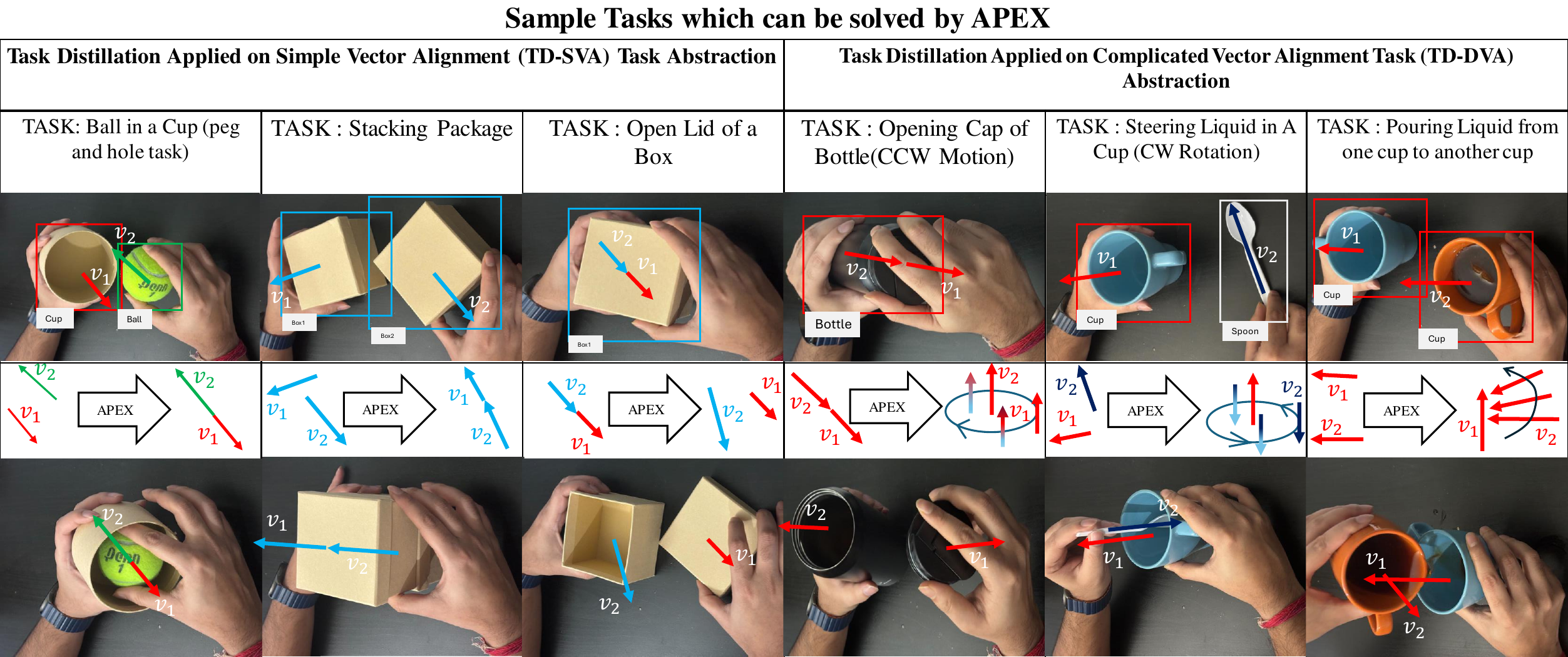}  
    \caption{Solving the real-world dexterous dual-arm robotic manipulation tasks. The first row indicates various robotic manipulation tasks. The second row is the initial position of the manipulation tasks. The third row distills the manipulation tasks by abstracting them as vector alignment problems, which can be solved using our proposed algorithm. The last row shows the successful completeness of the manipulation tasks.}%
    \label{fig: Methods task demos}                                   
\end{figure*} 

To date, diffusion models have emerged as a powerful tool in robotics, owing to their flexibility and multi-modality. However, their application in the context of dual-arm robotic manipulation remains relatively unexplored, particularly concerning real-time performance and the diversity of manipulation tasks.
In this paper, we propose a collision-free
diffusion model to address the aforementioned challenges. The key contributions
of our work are outlined below:

\begin{itemize}
\item
Abstracting real-life ambidextrous dual-arm robotic manipulation tasks as two vector alignment problems, a conceptualization that significantly simplifies the tasks and underscores the model's applicability across various real-world scenarios.
\item
Development of latent diffusion models for generating diverse robotic manipulation trajectories. Additionally, we integrate obstacle information using the classifier-guidance technique, ensuring the practicality and safety of the resulting manipulation trajectories.
\item
Validation of the proposed algorithm on a hardware platform featuring ambidextrous dual-arm robots. Our algorithm consistently produces successful and seamless trajectories across multiple tasks, outperforming conventional robotic motion planning algorithms. These findings carry implications for the future design of diffusion robots, particularly in tackling more complex robotic manipulation tasks effectively.
\end{itemize}

\section{Related work}
\subsection{Dual-arm robotic manipulation}


Dual-arm robotic manipulation has been traditionally formulated as a classical optimal control problem, which primarily involves minimizing a desired cost function using a simplified model of the dynamical system \cite{dual_arm_opt_control,
 He2022_dual_arm_opt_control}. Second, researchers also primarily relied on the classic sampling framework of Rapidly-exploring Random Trees (RRT)
algorithms \cite{dual_arm_rrt1, dual_arm_rrt2}, which mainly develops graph traversal methods in the robotic configuration space, whose trajectory planning ability becomes limited and useless if the dimension of the robotic configuration space becomes higher. Recently, reinforcement learning (RL) algorithms
\cite{dual_arm_robot_rl1, alles2022learning}, have been introduced by maximizing a reward function for accomplishing such dual-arm manipulation tasks. Imitation Learning (IL) algorithms initially gather a substantial number of task samples, such as human demonstrations, and then employs deep learning
architectures like convolutional/recurrent neural networks\cite{IL_dual_arm_manipulation, IL_dual_arm_2} or transformers\cite{KIM_IL_IROS_2021} to train the model through teacher-forcing training. However, the above learning-based methods may be severely suffered by the heterogeneous data distribution shift, i.e., the training and test data are sampled from different manipulation tasks or the obstacles distribution is biased, and limited trajectory generation diversity, which ultimately decreases their potential advantages. Although these approaches have proven highly effective in some cases, they do not address highly complex real-world working spaces with potentially dynamic obstacles or the full coordination of robotic arms and hands.

\subsection{Diffusion models}

Diffusion models have been illustrated with great potential as generative models
of images~\cite{diffusion_Model_survey} and videos~\cite{video_diffusion} and
outperform other generative models such as adversarial networks (GANs)
~\cite{classifier_guidance1} and variational autoencoders (VAEs). Ho et
al~\cite{ho2020denoising} proposes denoising diffusion probabilistic models
(DDPMs) by treating the training procedure as minimizing the mean-square error
of noise at each time step. To facilitate the sampling time, latent diffusion
models (LDMs) are proposed to first integrate autoencoders (AEs) for latent
representation and then to train the diffusion model in the corresponding latent
space~\cite{rombach2022high}. After sampling, the generated latent
representation will be sent through a decoder to recover the generated images.
For most of the diffusion models, the underlying backbone architecture is
selected as convolutional neural networks (CNNs), such as
U-Nets~\cite{ronneberger2015u}. To date, there are few works that incorporate the
diffusion models for robotic motion planning~\cite{dall_e_diff,
power2023sampling, janner2022diffuser, Carvalho_Motion_diffusion_2023_IROS,
yoneda2023diffusha}. For instance, \cite{dall_e_diff} utilized a zero-shot
learning mechanism with pre-trained generative models to complete object
arrangement tasks. On the contrary, \cite{janner2022diffuser} implemented a reward
mechanism along with diffusion model-based trajectory generation and the authors
proposed a reward-based guidance for optimization with maximum utility for task
completion.

\begin{figure*}                                     
    \centering                                             
    \includegraphics[width=\linewidth]{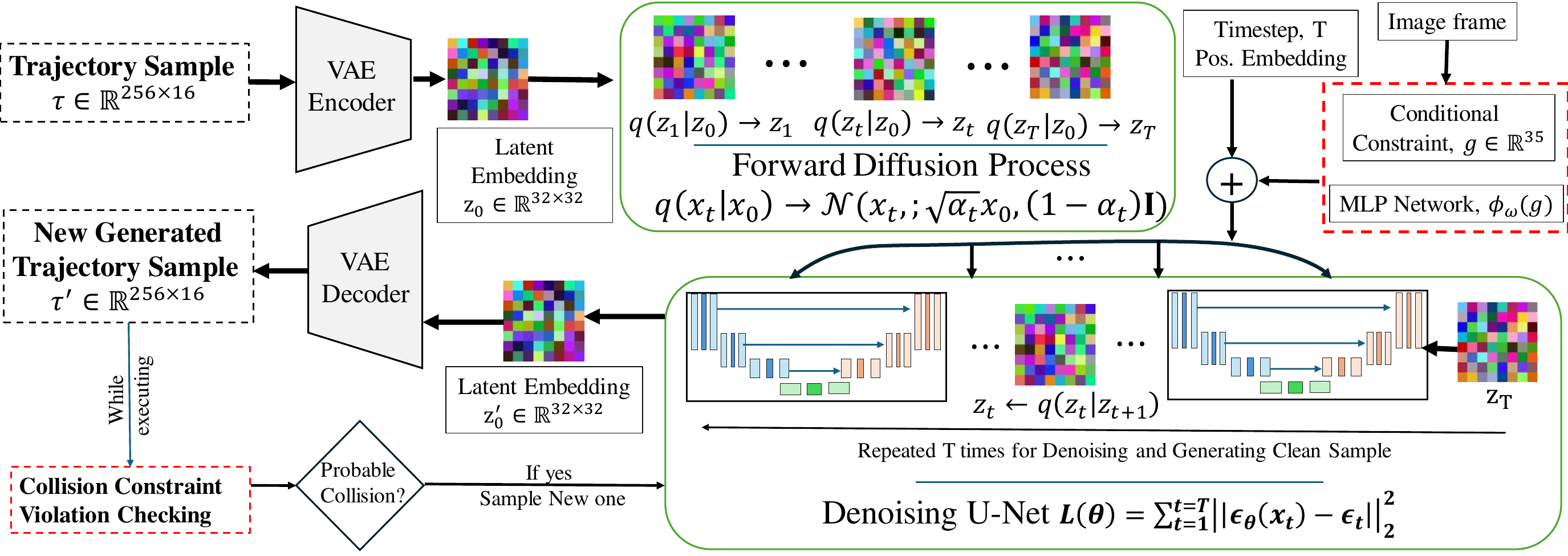}  
    \caption{The diagram of our ambidextrous dual-arm robotic manipulation algorithm using latent diffusion models to generate the collision-free trajectory. The input of our trajectory data $x$ first goes into a pre-trained variational autoencoder to have a low-dimensional latent embedding $z_0$, which will be used to train a diffusion model. In the sampling procedure, we randomly generate Gaussian noise trajectory $z_T$ and pass it through reverse denoising steps to generate clean trajectory latent embedding $z^{'}_0$, which can be decoded using a pre-trained VAE decoder. Further, to ensure the generated trajectory $x^{'}$ is collusion-free, we add two mechanisms on top of the latent diffusion models. First, we use the obstacle information $g$ as the conditional constraint in both training and sampling process. Second, our trajectory generation process works in a close-loop manner, where we check the possibility of collision violation and resample a new trajectory if necessary.}%
    \label{fig: Methods Collision-Free Generative Diffusion Models}                                   
\end{figure*} 

\section{Methods}
\subsection{Distill ambidextrous dual-arm manipulation tasks as vector alignment problems}

Many day-to-day tasks such as stacking packages or pouring liquid from one cup to another cup can be summarized and simplified. In our work, we distill the dual-arm manipulation tasks by vector alignment problems. Here, we demonstrated several dual-arm manipulation tasks as well as their distillations in Fig.\ref{fig: Methods task demos}. We first used the well-established YOLO algorithm (a typical type of convolutional neural network in computer vision) for object detection from the captured frame using the depth sensor~\cite{yolo_orientation}. The YOLO algorithm localized the object of interest at each time frame and detected the position and orientations of those objects. Then we drew simple 2D vectors $v_1$ and $v_2$ along the orientations of the objects as depicted in the upper row of Fig.\ref{fig: Methods task demos}. Once we captured the orientations of two vectors, we can convert the complex manipulation task to a simple vector alignment problem. Therefore, this distillation reduces computationally heavy image-based motion planning problems to a simple vector-conditioned trajectory planning problem. Take the first task (Ball in a cup) for an illustration, the two vectors are randomly placed initially, where we are required to develop a motion planning algorithm such that inital two vectors are aligned in the desired positions and opposite orientations. 
Once our proposed algorithm finishes the task, we can measure the robot configuration space using the measured initial state and final state for the performance evaluation of our algorithm. Further, we also consider dynamic obstacles during our tasks for the mimic of more complicated real-world robotic motion planning tasks.

Let $\mathcal{S}\subseteq \mathbb{R}^n$ and $U\subseteq \mathbb{R}^m$ define the state space and control input for our dual-arm robotic system. The robotic system evolves through the following discrete time-dynamics,
\begin{equation}
    \mathcal{S}_{t+1} = f(\mathcal{S}_t, u_t)
    \label{eq:system dynamics}
\end{equation}
where $f(\cdot)$ is a function controlling the discrete-time system progression for action $u_t$ from state $\mathcal{S}_t$ to $\mathcal{S}_{t+1}$. We assume the $\mathcal{S}_\text{safe}$ and $\mathcal{S}_\text{unsafe}$ are the collision-free and collision space with $\mathcal{S}_\text{safe} \cap \mathcal{S}_\text{unsafe} = \emptyset $, respectively. During the task distillation, we get the initial state vector, $\mathcal{S}_\text{ini}\in \mathcal{S}_\text{safe}$ and target vector, $\mathcal{S}_\text{goal}\in \mathcal{S}_\text{safe}$ and current point obstacle information. Together, all the above information constitutes a conditional guidance vector $g_t$. The goal of our algorithm is to generate a trajectory, $\tau=\{\tau_0,\cdots,\tau_{T-1}\}$ conditioned on $g_t$ such that it is safe against dynamically moving obstacles ($\tau \in \mathcal{S}_\text{safe}$) and drive the two-arm robotic system reaching the goal state $\mathcal{S}_\text{goal}$.

\subsection{Diffusion-based deep generative learning}
In this section, we develop the main component of our diffusion-based deep generative model for the synthesis of alignment trajectories. We consider the following Gaussian diffusion models, which include a forward process and a reverse process. The forward noise process describes conditional probability distribution $q(x_t|x_0)$ given the initial data distribution $q(x_0)$, which can be the collected trajectory data in our case. 
\begin{equation}
    \label{Methods: Gaussian diffusion models Forward}
    q(x_t|x_0) = \mathcal{N}(x_t;\sqrt{\Bar{\alpha}_t}x_0, (1-\Bar{\alpha}_t)\mathbf{I}),
\end{equation}
where $\Bar{\alpha}_t$ is are predefined constant for introducing noise at each forward process~\cite{ho2020denoising}. Using the property of Gaussian distribution and parameterization trick, one can directly sample $x_t$ using~\eqref{Methods: Gaussian diffusion models Forward} by $x_t = \sqrt{\Bar{\alpha}_t} x_0 + \sqrt{(1-\Bar{\alpha}_t)}\epsilon_t$, where $\epsilon_t$ follows standard Gaussian distribution $\mathcal{N}(\mathbf{0},\mathbf{I})$. The reverse diffusion process is to learn a set of the neural network parameters $\theta$ that reverse the forward noise process,  
\begin{equation}
    \label{Methods: Gaussian diffusion models Reverse}
    p(x_{t-1}|x_t) = \mathcal{N}(x_{t-1};\mu_\theta(x_t), \Sigma_\theta(x_t).
\end{equation}
Therefore, the goal of the diffusion models is to ensure the prediction $p_\theta(x_0)$ using learned neural network parameters $\theta$ that can recover the original data distribution $q(x_0)$ as close all possible, which can be quantified using evidence lower bound (ELBO)~\cite{ho2020denoising}. Thus, one can maximize the lower bound of the log-likelihood of $x_0$ with the following loss function 
\begin{equation}
\label{Methods: Gaussian diffusion models Complicated loss}
    L(\theta) = - \mathbb{E}_{x_{0:T \sim q}}[\text{log} p_\theta(x_{0:T}) - \text{log} q(x_{1:T}|x_0)].
\end{equation}
However, the above loss function~\eqref{Methods: Gaussian diffusion models Complicated loss} has been recognized as hard for fasting training since it focuses on image reconstructions. Essentially, the diffusion model need to care more about the reverse denoising process by focusing on cleaning the forward noise corruption, which inspires the design of noise prediction work~\cite{ho2020denoising},
\begin{equation}
\label{Methods: Gaussian diffusion models Simple loss}
    L(\theta) = \Sigma^{t=T}_{t=1}||\epsilon_\theta(x_t) - \epsilon_t||^{2}_{2}.
\end{equation}
Therefore, we utilize this simple mean-square error loss through our training process. In the sampling case, we randomly sample $x_T$ from standard Gaussian distribution $\mathcal{N}(\mathbf{0},\mathbf{I})$ and pass through the reverse denoising process to graduate clean the noisy trajectories step by step $p(x_{t-1}|x_t)$ (see Figure~\ref{fig: Methods Collision-Free Generative Diffusion Models}). The above equations~\eqref{Methods: Gaussian diffusion models Forward}, ~\eqref{Methods: Gaussian diffusion models Reverse},~\eqref{Methods: Gaussian diffusion models Simple loss} describe how we construct the diffusion model for the generation of new trajectories. Also, the backbone of the diffusion model follows the U-Net architecture, a typical type of symmetric conventional convolution neural networks (CNNs). 

However, the above naive diffusion model meets two difficulties for the robotic trajectory generations. First, the robotic planning trajectory usually has high temporal dimensions. Thus, directly training and sampling on the high-dimensional trajectory space is hard and time-consuming. Second, the manipulation of robotics usually involves the interaction of dynamic environments, meaning that the robot needs to react to the environmental obstacles and replan the trajectories to avoid any potential collisions. Therefore, we leverage the latent diffusion models by first training a variational autoencoder to embed the high-dimensional trajectory as $z = \text{Encoder}(\tau)$. Then, we train the proposed diffusion model in the embedded space. Last, we use a trained decoder $\tau = \text{Decoder}(z)$ to map the sampled low-dimensional trajectory embedding back to the original trajectories for later manipulation of ambidextrous dual-arm robots. To deal with the second difficulty, we integrate the classifier guidance technique as the real-time feedback condition to guide the generation of safe and feasible trajectories (see next section).

\subsection{Obstacle guidance as feedback to close the loop}
\label{Methods: Classifier guidance as feedback to avoid obstacles and realize alignment}
The most advanced diffusion robot works mainly applied the diffusion framework for synthesizing planning trajectories without considering dynamic environmental obstacles ~\cite{janner2022diffuser, Carvalho_Motion_diffusion_2023_IROS, Chi2023DiffusionPV, reuss2023goal}. Our second innovative contribution is to use the classifier guidance technique~\cite{classifier_guidance1, ho2022classifier} as the real-time feedback condition to guide the generation of safe and feasible trajectories. Specifically, we assume at certain steps, our camera sensor can take the image of the current interaction of the robotic arms and environment (denoted as $g_t$), abstractly representing the current alignment procedure and potential environmental obstacles. The conditional diffusion models using the obstacle guidance technique can be developed under the Bayes theorem $p(x|g)\propto p(x)p(g|x)$,    
\begin{equation}
\label{Methods: Gaussian diffusion models Classifier Guidance Score function}   
\nabla_x \text{log} p_\theta(x|g) =  \nabla_x \text{log} p_\theta(x) + \gamma \nabla_x \text{log} p_\theta(g|x),
\end{equation}  
where $\gamma$ controls the guidance scale; $g$ is the interaction information that we described above. Put equation~\eqref{Methods: Gaussian diffusion models Classifier Guidance Score function} into the diffusion model, it turns out that we only need to modify noise prediction at the sampling stage via
\begin{equation}
\label{Methods: Gaussian diffusion models modified noise prediction}
   \hat{\epsilon}_\theta(x_t) = \epsilon_\theta(x_t)  - \gamma \sqrt{(1-\alpha_t)}\nabla_{x_t} \text{log} p_\theta(g|x).
\end{equation}
In the extreme case $\gamma = 0$, this conditional diffusion model degrades to the previous diffusion model, which can generate diverse but unsafe manipulation trajectories. On the contrary, we impose the conditional generation of collision-free trajectory as we increase the guidance scale $\gamma$. Therefore, by integrating the classifier-guidance technique as obstacle-guidance, we essentially build a closed-loop trajectory generation procedure, where we adapt the prior generated trajectory at certain time steps using the real-time obstacle feedback.     

To this end, we complete the development of our proposed algorithm. A summary of the algorithm can be found in the algorithm~\ref{Algorithm 1} and figure~\ref{fig: Methods Collision-Free Generative Diffusion Models}.
\begin{algorithm}
	\caption{APEX}
	\label{Algorithm 1}
        \textbf{--------TRAINING--------}\\
        \small
		\textbf{Input:} Dataset, $\mathcal{D}:\{\text{Trajectory} \;\tau_i, \text{Guidance} \;g_i\}_{i=1}^N$, Conditional Diffusion Model, $\epsilon_\theta$; $\Bar{\alpha}_t, T$ \\
        \textbf{Output:} Trained model, $\Tilde{\epsilon_\theta}$\\
  \SetKwProg{Fn}{Function}{}{}
\Fn {\text{train($\mathcal{D}, \Bar{\alpha}_t, \epsilon_\theta, T$):}}{
		\While{not converged}{
        \scriptsize \texttt{\# sample batch of traj. and respective conditional vector}\\
        \small
        $(\tau_0,g_0)\sim\mathcal{D}, \epsilon\sim\mathcal{N}(0,I), t\sim\mathcal{U}(1,T)$\\
        \scriptsize \texttt{\# concat time-embedding and $g_i$}\\
        \small
        $t\_cond = concat(pos\_emb(t),\phi_\omega(g_i))$\\
        \scriptsize \texttt{\# create noisy trajectory}\\
        \small
        $\tau_\epsilon(t)\leftarrow\sqrt{\Bar{\alpha}_t(t)}\tau_0+\sqrt{1-\Bar{\alpha}_t(t)}\epsilon$ \\
        \scriptsize \texttt{\# MSE Error calc.}\\
        \small
        $L_\theta=||\epsilon-\epsilon_\theta(\tau_\epsilon(t), t\_cond)||_2^2$ \\
        \scriptsize \texttt{\# Gradient Update}\\
        \small
        $\theta\leftarrow\theta+\nabla_\theta L_\theta$
        }
}
\normalsize
\textbf{--------SAMPLING--------}\\
\small
\textbf{Input:} Frame $f_0$, Trained Model, $\Tilde{\epsilon}_\theta$ , $\lambda_{obs}$\\
        \textbf{Output:} Feasible Trajectory, $\tau^*$\\
  \SetKwProg{Fn}{Function}{}{}
\Fn {\text{execute($f_0,\Tilde{\epsilon}_\theta,\lambda_{obs})$:}}{
    \scriptsize \texttt{\# Task Distil. for Vector Alignment and obs info}\\
    \small
    $g_0\leftarrow vect\_align(f_0)$ \\
    \scriptsize \texttt{\# Sample Noisy Trajectory }\\
    \small
    $\tau_T\leftarrow \mathcal{N}(0,I)$ \\
    \scriptsize \texttt{\# Noise Prediction}\\
    \small
    $\epsilon_T\leftarrow\epsilon_\theta(\tau_T,concat(pos\_emb(T),\phi_\omega(g_0)))$ \\
    \scriptsize \texttt{\# Initial Safe Trajectory Calc.}\\
    \small
    $\tau_0\leftarrow traj\_calc(\alpha_t,\epsilon_T,\tau_T)$ \\
    \While{not done}{
        \texttt{CALCULATE:}\\
        Distance, $d_{obs}\gets GJK(robot,obstacle)$ \\
        \scriptsize \texttt{\# Collision Checking}\\
        \small 
        \eIf{$d_{obs}<\lambda_{obs}$}{
        \texttt{UPDATE:}\\
        \scriptsize \texttt{\# Sample new trajectory for collision avoidance}\\
        \small
        $\tau_{new}\leftarrow generate\_traj(f_{curr},\Tilde{\epsilon}_\theta)$
        }
        { \texttt{EXECUTE:}\\
        \scriptsize \texttt{\# Action Execution following generated traj.}\\
    \small
        Dual Arm-hand Control, $\tau_i\rightarrow \{[\theta_0^k,\cdots\theta_6^k]_{k=1,2}, \theta_h^1,\theta_h^2\}_{i}$  
    }

}
}
\end{algorithm}
\subsection{Platforms and implementation details}
We utilized two high-fidelity 7-DoF Franka Emika Panda robot arms and assembled them on two separate table-top for smoother and collision-free operation. We used two anthropomorphic five-fingered hands named QB SoftHand2 Research in left-right hand combinations. 
To locate objects and track dynamic obstacles with depth information, we used an Intel RealSense Depth Camera D435i. On each frame captured by depth camera, we run YOLO for object detection and obstacle localization. Later, we detected the orientation and position of vectors associated with each object. To differentiate the obstacle from objects, we used color-based segmentation on detected frames. By applying the conventional hand-to-eye calibration, we transported the $[x,y,z]_o$ information from the camera reference frame to the robot coordinate system. Our algorithm utilized this feedback from depth sensors to avoid obstacles in real-world scenarios by computing the distance between the 3D geometry of the obstacle and the collision meshes of robot links using the Gilbert–Johnson–Keerthi distance (GJK)\cite{Khan2020} algorithm. This distance feedback triggers the controller to search for a new trajectory if chances of collision occur while completing the vector alignment task. To validate our approach, we replicated the exact model of the Franka Emika Panda Arms with QB SoftHand2 in the Robot Operating System (ROS). We utilized the libfranka and Franka ROS, to establish low-latency and low-noise communication protocols for data processing and parallel execution among simulation environments and real hardware. QB Softhand2 research hands were operated through serial communication in parallel for object grasping and placement into the assigned alignment by associated vectors.
\begin{figure*}                                     
    \centering                                             
    \includegraphics[width=\linewidth]{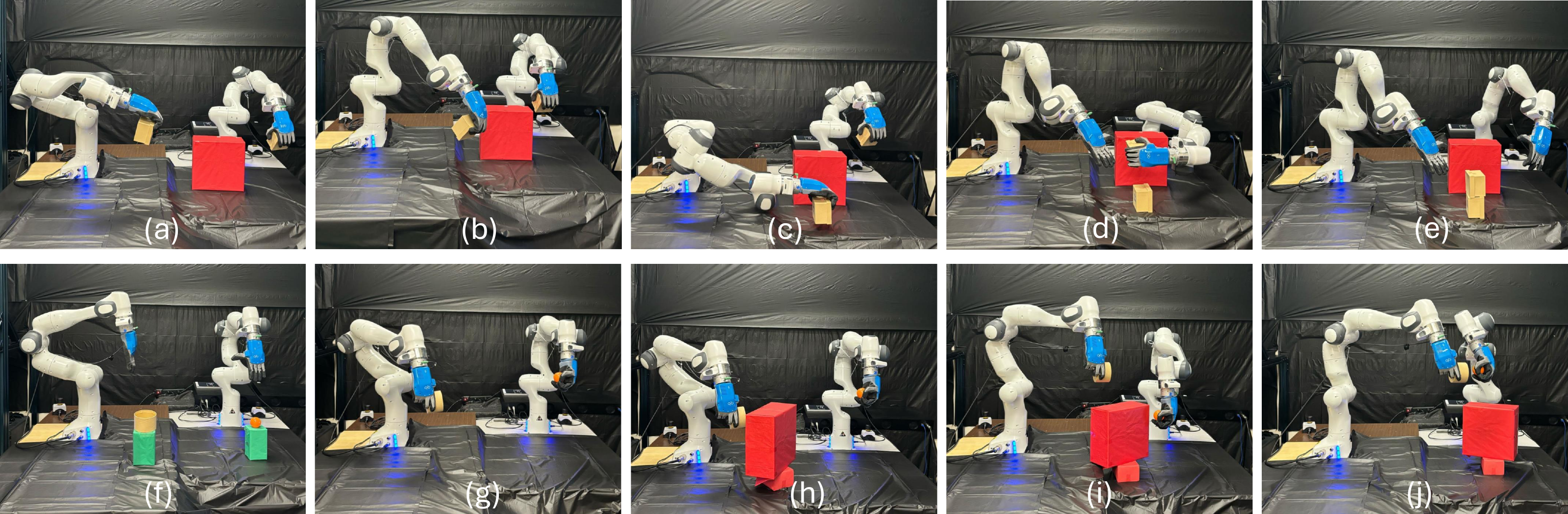}  
    \caption{Snapshots of hardware demonstration from vertical (a)-(e) and horizontal alignment task (f)-(j). (a)-(b) Robotic arms grasp the boxes from a table. (c)-(d) Right robotic arm avoids possible collision (red box). (e) Successful task completed. (f)-(g) Two arms take the cup and the ball to prepare a task in hand. (h) A dynamic obstacle appears in the workspace and the left robotic arm faces probable collision, (i) The Left arm moves over the obstacle to avoid collision. Now, the right arm gets into a probable collision with the moving red box. (j) Both arms dodge the obstacles and complete the task.}%
    \label{fig: hardware task}                                   
\end{figure*} 

\begin{figure*}                                     
    \centering               
    \includegraphics[width=\linewidth]{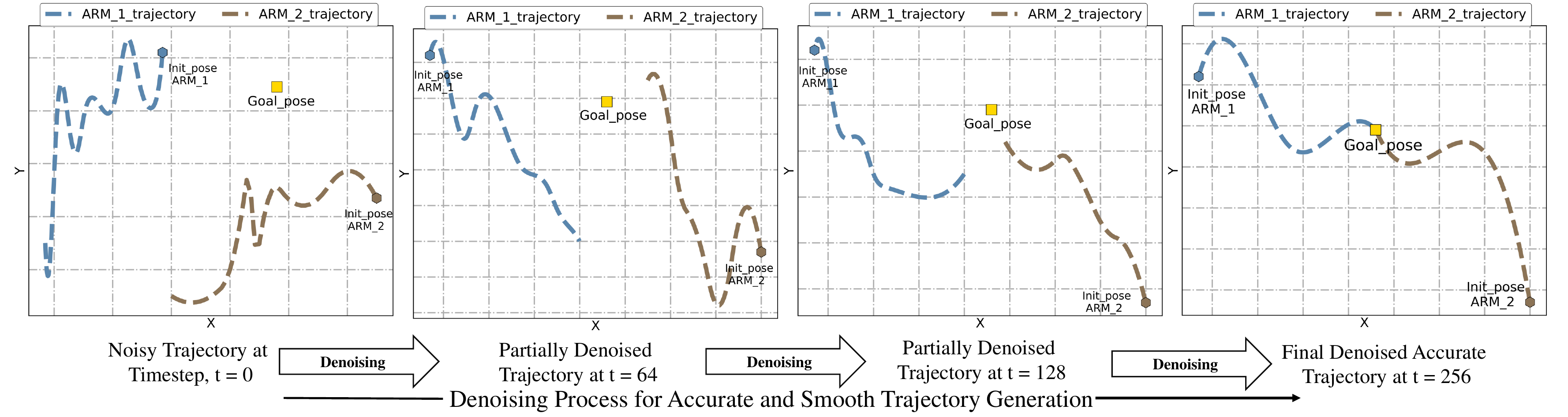}  
    \caption{Denoising diffusion process for the generation of manipulation trajectories. From left to right, we gradually show how the denoising diffusion process generates a smooth and feasible trajectory to reach the predefined goal position.}%
    \label{fig: diffusion step}                                   
\end{figure*}

We collected high-dimensional trajectory $\tau_i$ samples for dual-arm manipulation setup from ROS, where each data sample $\tau_i^j$ contains total $16$ float values -- $14$ joint values from two robotic arms $[\theta_0^k,\cdots\theta_6^k]_{k=1,2}$ and $2$ synergy joint values from two robotic hands $[\theta_h^1,\theta_h^2]$. We only use robotic hands for grasping tasks. Thus, we only exploit the first synergy of robotic hands. As a result, each robotic hand can either close or open, where the synergy value indirectly controls the $19$ self-healing finger joints. The conditional constraint vector has $35$ float values -- $16$ values for initial or current system state which contains $\{[\theta_0^k,\cdots\theta_6^k]_{k=1,2}, \theta_h^1,\theta_h^2\}_{ini}$, $16$ values for final or target state $\{[\theta_0^k,\cdots\theta_6^k]_{k=1,2}, \theta_h^1,\theta_h^2\}_{tar}$, $3$ values for obstacle point $[x,y,z]_o$. Then, the conditional vector $g$ goes through a MLP network $\phi_\omega(g)$ parameterized by $\omega$ and the output of $\phi_\omega(\cdot)$ applies residual connections to each convolution layer in the U-Net(see Fig.~\ref{fig: Methods Collision-Free Generative Diffusion Models}), and later also uses positional embedding of each timestep during the model training. After obstacle detection, the GJK algorithm localizes the nearest point on obstacle body by calculating the distance from full robot meshes. We collected 500k collision-free trajectory samples with different starting and ending alignment configurations from our simulation platform. Each trajectory samples consist $256$ discrete trajectory steps, which are from the starting state to the final ending state. Also, the dense trajectory samples allow us to perform adaptive planning from any initial state to the target goal state and handle dynamic obstacles simultaneously. Our algorithm was implemented using the Pytorch library and ran on a Lambda QUAD GPU workstation equipped with an Intel Core-i9-9820X processor and $4$ Nvidia $3080-ti$ GPU machines. It took 10 hours to train our collision-free generative diffusion models.      

\subsection{Experiments setup and evaluation metrics}
We design both comprehensive Monte Carlo computer simulations and hardware experiments to evaluate our proposed APEX algorithm for achieving dual-arm alignment manipulations. Our experiments mainly focus on two aspects: A) the vertical alignment, such as stacking, which can simplify the package sorting and stacking tasks in real life; B) the horizontal alignment, such as peg and hole task, which can be used to represent any insertion task in real life. We believe that by illustrating the performance on the above vertical and horizontal alignment tasks, we essentially evaluate our algorithms across a wide range of dual-arm robotic manipulation tasks. Further, we also implement typical existing robotic motion planning algorithms in our experiments, where we include the RRT~\cite{dynamicrrt} and GPMP~\cite{mukadam_GPMP_ICRA_2016}, a transformer-based imitation learning algorithm for motion planning~\cite{KIM_IL_IROS_2021} and our prior VAE-based graph traversal planning algorihtm~\cite{Dastider_Damon_IROS_2023}. For object grasping, we utilize advanced functionalities from ROS MoveIt Grasps API.


We define the following evaluation metrics to effectively quantify the performance. First, we define the failure ratio, 
\begin{equation}
    R_f = \frac{F}{N} \times 100\%,
\end{equation}
where $F$ counts the failed tasks and $N$ denotes the total number of tasks that we performed. Therefore, smaller $R_f$ means the generated trajectories successfully avoid obstacles and realize better alignment performances. 
Then, we quantify the smoothness of the executing trajectory of dual-arm robots by summing up the joint space configuration change in sampled trajectory as
\begin{equation}
    S_r = \sum_{i=0}^{N-1}||\Theta_i - \Theta_{i+1}||_2
\end{equation}
where $\Theta_i$ is joint space configuration at $\tau_i$.
Last, we use the Frechet inception distance (FID) to measure the generation diversity of our obstacle-guidance diffusion models,
\begin{equation}
    \text{FID} = ||\mu_R - \mu_G||^2 + \text{Tr}(\Sigma_R +\Sigma_G - 2(\Sigma_R\Sigma_G)^{\frac{1}{2}}),
\end{equation}
where $\mu_R $ and $\mu_G $ denote the feature-wise mean of the oracle and generated trajectories; $\Sigma_R $ and $\Sigma_G $
are the covariance matrix for the real and generated feature trajectories; $\text{Tr}(.)$ refers to the trace operation in linear algebra. Notice that researchers widely use the above FID score to quantify the realism and diversity in evaluating the performance of generative models\cite{FID_2017}. Here, we focus on the trajectory data and use pre-trained VAE model on our collected dataset. Therefore, we use the mean $\mu_G$ and variance $\Sigma_G$ calculated from the VAE model for the measurement of the FID score.

\section{Results}
We use comprehensive experiments to evaluate our proposed APEX algorithm. Specifically, our experiments seek to investigate the following questions: 
\begin{enumerate}
\item Can the ambidextrous dual-arm robots successfully complete the manipulation tasks using APEX?
\item What are the qualitative advantages and the essential trade-off of APEX compared to SOTA algorithms?  
\item Why the obstacle guidance and closed-loop fashion are important for safe planning?  
\item Are the trajectories generated by our algorithm diverse?
\end{enumerate}

\subsection{Successful and smooth ambidextrous dual-arm robotic manipulation trajectory}
We first presented horizontal and vertical task demos to visualize the trajectories generated by our proposed algorithm. Fig.~\ref{fig: hardware task} presented the successful ambidextrous dual-arm robotic manipulation performance in both vertical and horizontal tasks. Here, we took the horizontal tasks for example. We observed that the two arm-hand robots were far away and randomly placed in the configuration space at the beginning (see Fig.~\ref{fig: hardware task}(f)). After our diffusion model generated a trajectory, two robots gradually began to align resulting in a smooth trajectory (see Fig.~\ref{fig: hardware task}(g)). During the middle point of the alignment, an obstacle was imposed manually, which may resulted in the potential collision (see Fig.~\ref{fig: hardware task}(h)-(i)). Therefore, our robots stopped for avoiding any collision and generated a new trajectory, which lifted the two arms up to avoid the imposed hard obstacles (see Fig.~\ref{fig: hardware task}(h)-(i)). Finally, our two arm-hand robots successfully realized horizontal alignment, which completed the manipulation task ((see Fig.~\ref{fig: hardware task}(j)). Similar successful and smooth ambidextrous dual-arm robotic manipulation trajectories are observed in our vertical tasks (see Fig.~\ref{fig: hardware task}(a)-(e)).

Meanwhile, we also looked into the denoising diffusion process in our APEX algorithm. The beginning manipulation trajectory ($t = 0$) was randomly sampled as a Gaussian image, which led to a rigid and aimless trajectory. Then, the denoising diffusion process gradually cleaned the noise trajectory (snapshots at $t = 64$ and $t = 128$). Finally, the cleaned trajectory enabled a successful completion of the task and the planar movement of the arm-hand robots was smooth ($t = 256$). The above two illustrations of manipulation tasks provide interesting collision-free trajectory generation results and show promising advantages of our proposed generative diffusion model. 

\subsection{Comparison results with SOTA algorithms}

Next, we comprehensively evaluated our proposed algorithms for achieving various vertical and horizontal tasks from different initial positions and orientations. Further, to illustrate the potential advantages of our proposed algorithm, we compare our algorithm with other SOTA motion planning algorithms to evaluate the potential advantages.
\begin{figure}                                     
    \centering                                             
    \includegraphics[width=\linewidth]{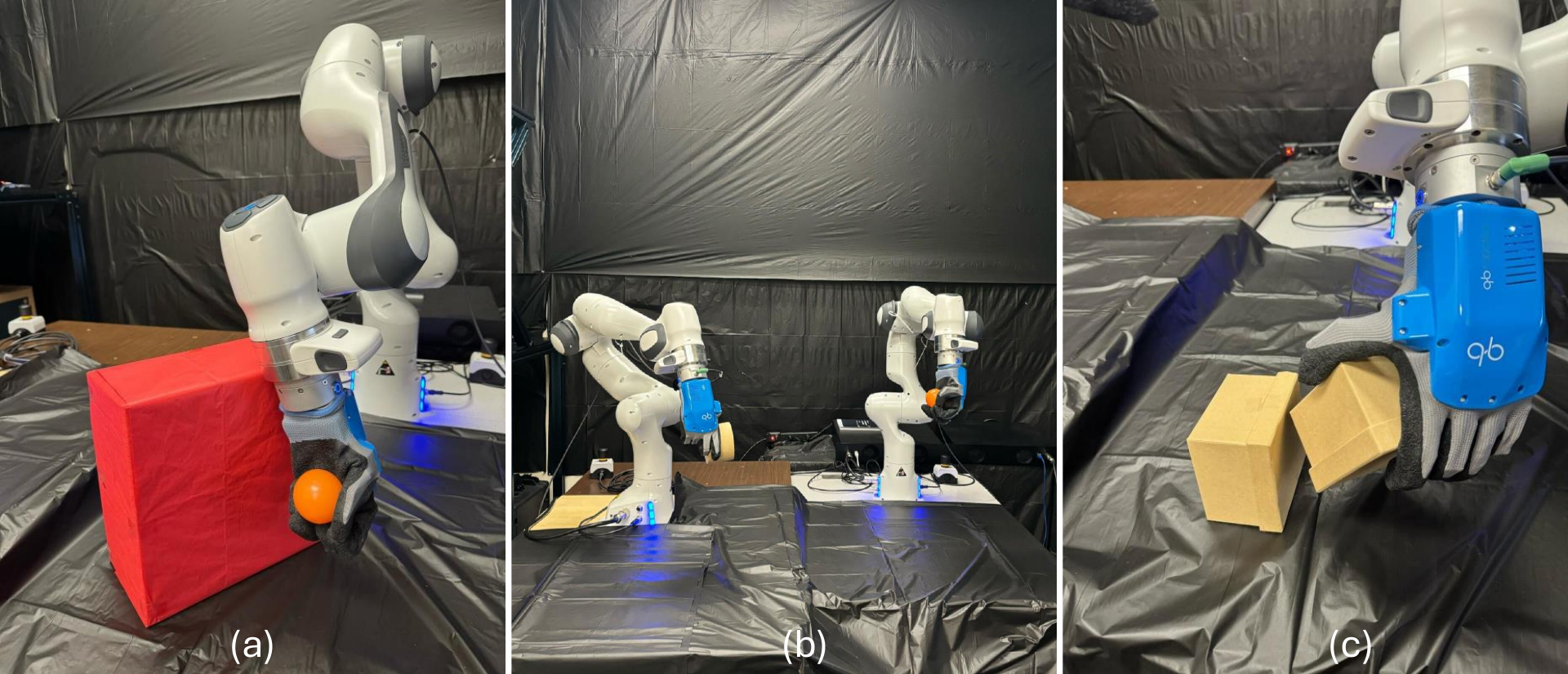}  
    \caption{
        (a) Collision While doing horizontal alignment task, (b) Two arms are stuck in the middle and can not replan, (c) Incorrect orientation position for vertical aloignment task}%
    \label{fig:failure}                                   
\end{figure}

\begin{figure}                                     
    \centering                                             
    \includegraphics[width=\linewidth]{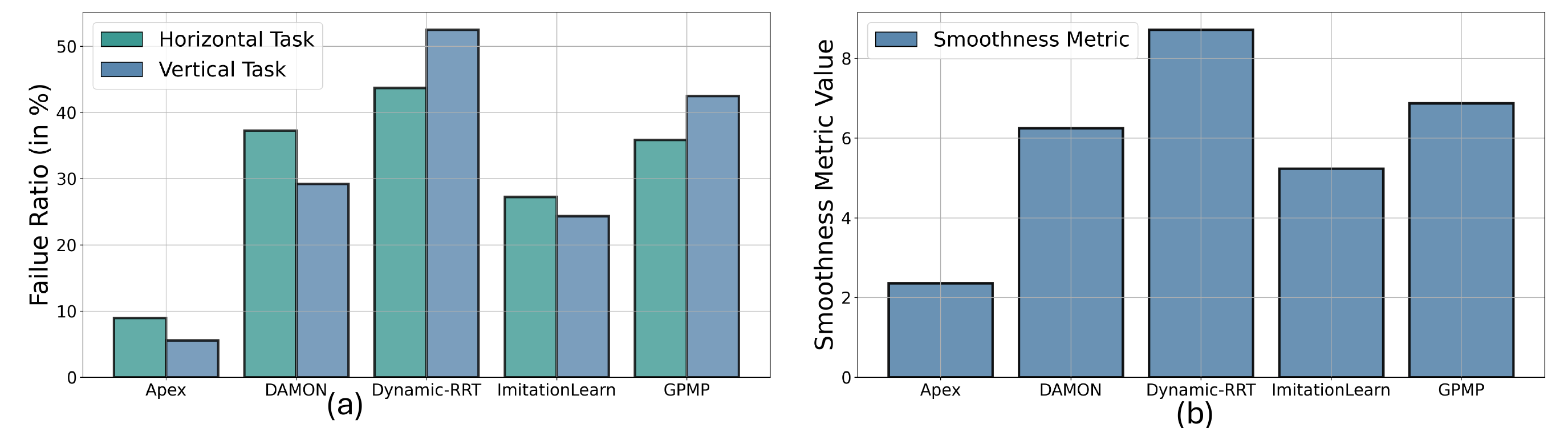}  
    \caption{
        (a) Comparison of failure ratio with SOTA algorithms, (b) Comparison of the smoothness with SOTA algorithms.}%
    \label{fig:comp_SOTA}                                   
\end{figure} 

We started by looking some failure cases of both vertical and horizontal alignments using other algorithms. Fig.~\ref{fig:failure} provides various failure examples. For example, Fig.~\ref{fig:failure}(b) showed that the traditional RRT seemed simply generated half of the trajectories, where two robotic arms stopped during the alignment; Fig.~\ref{fig:failure}(a) showed that the imitation learning algorithm can not deal with the obstacles, leading to a collision; Fig.~\ref{fig:failure}(c) showed that how DAMON failed to complete the alignment task and ended in a faulty configuration for box placement task. In total, fig.~\ref{fig:comp_SOTA}(a) gave the comprehensive comparison of different algorithms in executing vertical and horizontal tasks. We observed that our prior work DAMON and the most advanced imitation learning (IL) algorithms achieved better performance compared to the RRT and GPMP (Failure ratio: DAMON 32.65 v.s. IL 25.79 v.s. RRT 48.12 v.s. GPMP 39.26). However, our proposed diffusion-based generative models achieved extraordinary performance with the lowest failure ratio across both vertical and horizontal tasks (Failure ratio (avg): vertical 5.57 and horizontal 8.95). We argue that the possible reason why the vertical failure ratio is even lower than the horizontal task is due to the more complicated orientation alignment in horizontal tasks. For example, the vertical alignment task mimics the stacking manipulation, which does not involve orientation manipulations, whereas horizontal alignment tasks such as ball-insertion-in-a-cup require more complicated orientational manipulations.
Above all, our APEX algorithm achieved the best alignment performance across all comparison algorithms and tasks.

\begin{figure*}                                     
    \centering                 \includegraphics[width=\linewidth]{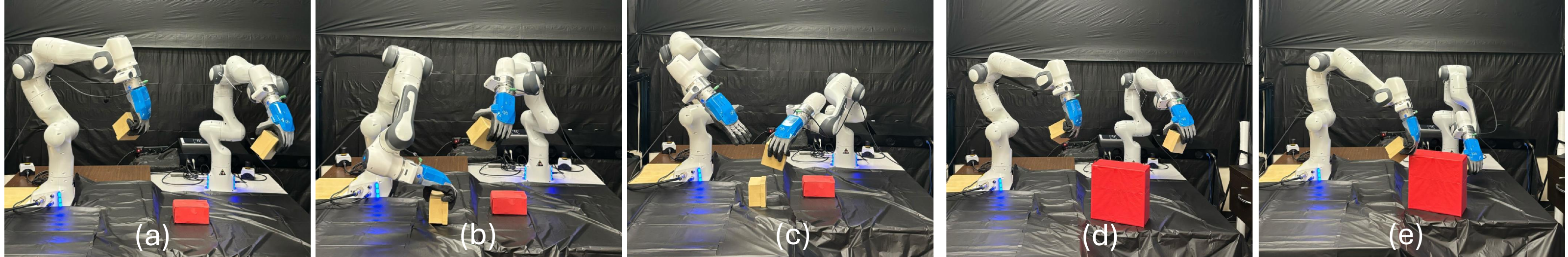}  
    \caption{Snapshots of hardware demonstration. (a)-(c) Without guidance, the robots can only operate under small obstacles and partially complete the task. (d)-(e)The right robotic arm collides with a larger obstacle can not complete the task.}%
    \label{fig: guidance free hardware task}                                   
\end{figure*}

Second, smoothness is also very important for avoiding awkward robotic manipulation behaviors. Prior demos illustrate that our diffusion models can generate smooth trajectories. One possible mechanism is the reverse diffusion process, where we use the trained noise prediction network to gradually denoise from a pure Gaussian noise trajectory. Advanced imitation learning (IL) algorithms also have mechanisms to ensure the smoothness of trajectory since the goal of IL is to mimic the training trajectories with mean square error loss. Fig.~\ref{fig:comp_SOTA}(b) gave the overall smoothness metric in our experiments, where we observed that our diffusion models achieved the smallest value indicating the most smooth manipulation trajectories ($S_r$ ours: 2.36 v.s. DAMON 6.25 v.s. IL 5.23 v.s. RRT 8.72 v.s. GPMP 6.87).    


In total, the above experimental results showed that our proposed diffusion models have many promising advantages such as lower failure rate and smooth trajectory, and outperformed other algorithms.

\subsection{Importance of obstacle guidance as real-time feedback to avoid obstacles and achieve alignment}

In Methods section~\ref{Methods: Classifier guidance as feedback to avoid obstacles and realize alignment}, we emphasized that integrating the obstacle guidance technique is important as it functions as the real-time closed-loop feedback condition to guide the generation of safe and feasible trajectories. Therefore, we use an ablation study to comprehensively evaluate the functionality of the obstacle guidance component.

\begin{figure}                                     
    \centering                                             
    \includegraphics[width=\linewidth]{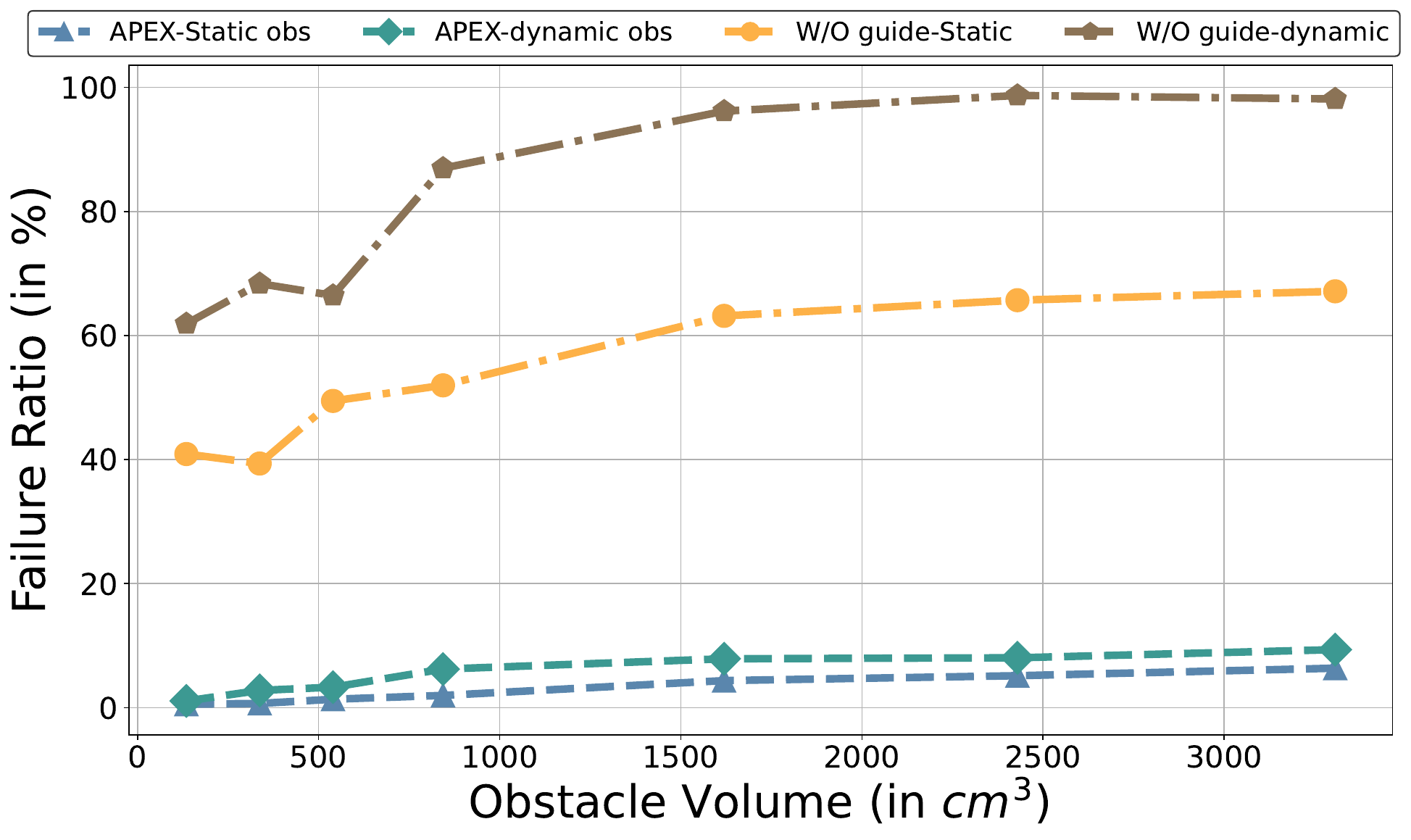}  
    \caption{
        Comparison of Failure Ratio with APEX and guidance-free stable diffusion algorithm for static and dynamic obstacle}%
    \label{fig:comp_guidance_free}                                   
\end{figure} 

Similar as the prior section, we first provide the manipulation demos for dynamical obstacles. Here we focused on two algorithms, our proposed algorithm and our algorithm without the obstacle guidance component. From the demos (see Fig.~\ref{fig: guidance free hardware task}(a)-(c)), we observed that if the obstacles were small, it was still possible for the no-guidance diffusion model to generate a safe and feasible manipulation trajectory. It can be understood if the sampler was lucky enough since the obstacles only occupied a small area in the configuration space. However, when we increased the size and number of obstacles or even enabled dynamical obstacles, the chance level using the diffusion model without guidance dropped drastically, leading to the collision trajectories (see Figure~\ref{fig: guidance free hardware task}(e)). On the contrary, our obstacle-guidance diffusion models can incorporate real-time video sensor images as the conditional guidance for the generation of future potential collision-free trajectories, which consistently resulted in lower failure ratios even in dynamical obstacle cases. We summarized the experimental results in Fig.~\ref{fig:comp_guidance_free}, where our methods achieved a significantly lower failure ratio compared to the no-guidance diffusion model (Failure ratio in static for the largest obstacle: ours 6.56 v.s. no-guidance diffusion model 67.19; Failure ratio in dynamic and large obstacle: ours 9.36 v.s. no-guidance diffusion model 98.45). We further showed in Fig.~\ref{fig:comp_guidance_free} that the failure ratio for the no-guidence diffusion model increased as the obstacle volume increased. On the contrary, APEX showed consistently low failure ratio across static and dynamic obstacles. Thus, our ablation study confirmed the essential contribution of the obstacle guidance technique in generating safe and feasible ambidextrous dual-arm robotic manipulation trajectories.

\subsection{Diversity of the trajectory generation}

\begin{figure}                                     
    \centering              
    \includegraphics[width=\linewidth]{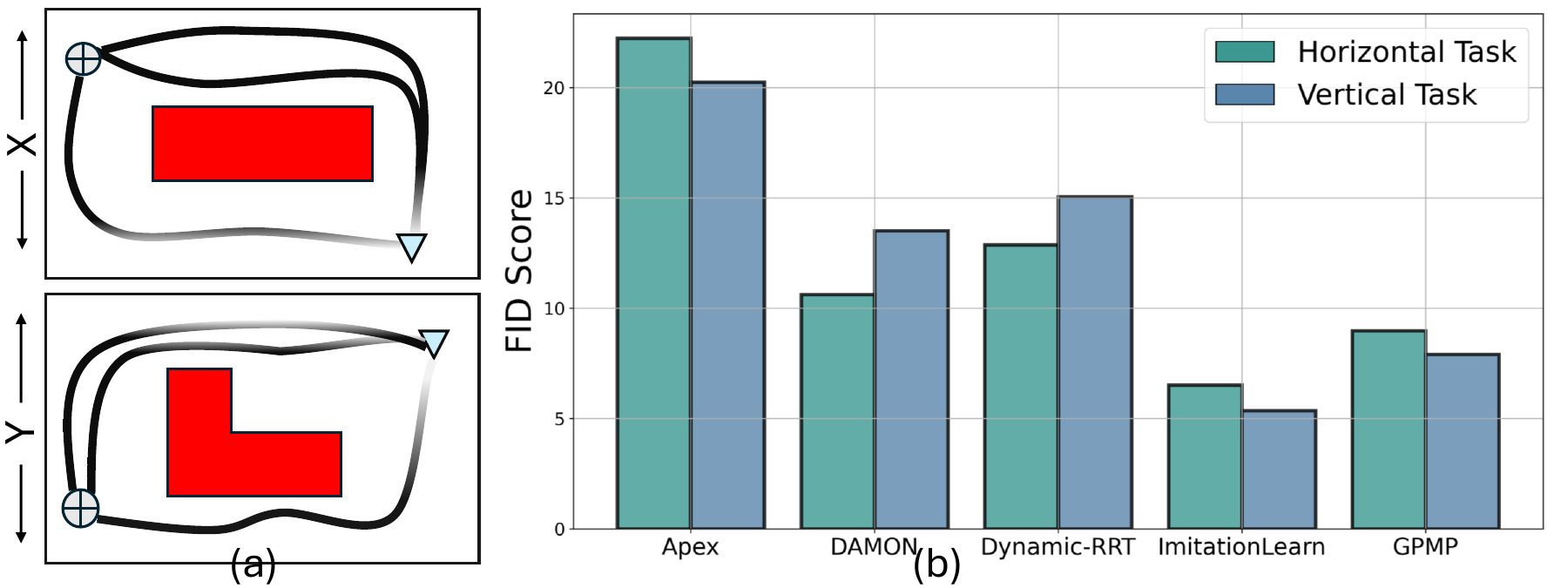}  
    \caption{
        (a) Several collision-free trajectories are generated from APEX. The $\bigoplus$ denotes the starting position and $\bigtriangledown$ denotes the goal position. The upper and lower panels are in X and Y coordinations. (b) FID comparison.}%
    \label{fig:FID_SOTA}                                   
\end{figure}

Last, we show two interesting findings. It is believed that by using different initial Gaussian noise, the diffusion models can generate various types of images or videos. Therefore, we also test our obstacle-guidance diffusion models by giving different initial Gaussian noise in the sampling process. We observed that our models indeed can generate multiple feasible and collision-free manipulation trajectories (see Fig.~\ref{fig:FID_SOTA}(a)), meaning that our model essentially learned complicated manipulation-task dependent distributions and realized effective multi-modal sampling from the learned distribution. Compared with other algorithms, our methods still have the most diverse trajectory generation performance. Further, we compute the FID for measurement of the generation diversity. Our FID results also confirmed that our proposed obstacle-guidance diffusion achieved the highest score across all the comparison algorithms (see Fig~\ref{fig:FID_SOTA}(b)). Second, regarding to the FID score comparison between our obstacle-guidance model and no-guidance diffusion model. We observed that our obstacle-guidance model has the lower FID score but faster trajectory generation speed (the FID score: ours 21.51 v.s. no-guidance diffusion model 48.63; the computational time ours 8.36s v.s. no-guidance diffusion model 19.63s), which revealed another trade-off that we sacrifice part of the generation diversity but both improve the sampling speed and ensure the feasibility and safety of the generated trajectories.   

\section{Discussion and Conclusion}

Effectively coordinating dual-arm movements while controlling 5-finger robotic hands to perform manual tasks poses serious challenges for real-time execution in a cluttered environment. APEX successfully leverages deep generative models for robotic motion planning to synthesize diverse feasible trajectories precisely and accurately. This achievement leads to low failure rates and flexible planning trajectories, fundamentally differing from traditional robotic planning algorithms and prior imitation learning-based deep neural networks.

APEX introduces three major innovations. Firstly, our task distillation dramatically simplifies training costs and enhances algorithm efficiency, thus facilitating adaptive obstacle avoidance, which proves crucial in addressing dynamic obstacles.
Secondly, the task demonstrations of APEX using latent diffusion models~\cite{rombach2022high},
mainly consisting of a Variational Autoencoder (VAE) and conventional U-Net structure,
were trained and evaluated separately as the initial trial of our algorithms.
Lastly, to further ensure that the generated trajectory is feasible and collision-free, we incorporate the obstacle-guidance technique by utilizing real-time obstacle and target information to construct conditional diffusion models.


APEX exhibits the capability to generate diverse feasible trajectories with lower failure rates for completing both vertical and horizontal dual-arm robotic manipulation tasks compared to state-of-the-art approaches. We validate the effectiveness of our proposed algorithm through computer simulations and experiments with ambidextrous dual-arm robotic systems. These results offer promising prospects for the future development of ambidextrous dual-arm robotics to tackle more intricate tasks. Given the generalizability of our proposed algorithm, we also foresee future explorations in other robotic systems such as robotic dogs~\cite{zhao_legged_robot}.

\bibliographystyle{./bibliography/IEEEtran}
\bibliography{bib}

\end{document}